\documentclass{article}

\usepackage[preprint]{neurips_2019}

\usepackage[utf8]{inputenc} 
\usepackage[T1]{fontenc}    
\usepackage{hyperref}       
\usepackage{url}            
\usepackage{booktabs}       
\usepackage{amsfonts}       
\usepackage{nicefrac}       
\usepackage{microtype}      
\usepackage{verbatim}
\usepackage{epsfig}
\usepackage{epsf}
\usepackage{epstopdf}
\usepackage{graphicx}
\usepackage{subfigure}
\usepackage{amsmath}
\usepackage{bigints}

\title{Tutorial: Deriving the Standard Variational Autoencoder (VAE) Loss Function}

\author{%
  Stephen G. Odaibo\thanks{Correspondence:  stephen.odaibo@retina-ai.com} \\
  (1) Department of Machine Learning Research\\RETINA-AI Health, Inc.\\
  (2) Department of Head \& Neck Surgery\\ Ophthalmology Section\\ MD Anderson Cancer Center\\
  \texttt{stephen.odaibo@retina-ai.com} \\
}

\begin{document}

\maketitle

\begin{abstract}
In Bayesian machine learning, the posterior distribution is typically computationally intractable, hence variational inference is often required. In this approach, an evidence lower bound on the log likelihood of data is maximized during training. Variational Autoencoders (VAE) are one important example where variational inference is utilized. In this tutorial, we derive the variational lower bound loss function of the standard variational autoencoder. We do so in the instance of a gaussian latent prior and gaussian approximate posterior, under which assumptions the Kullback-Leibler term in the variational lower bound has a closed form solution. We derive essentially everything we use along the way; everything from Bayes' theorem to the Kullback-Leibler divergence.
\end{abstract}

\section*{Bayes Theorem}
 Bayes theorem is a way to update one's belief as new evidence comes into view. The probability of a hypothesis, $z$, given some new data $x$, is denoted, $p(z|x)$, and is given by

  \begin{equation}
  p(z|x)=\frac{p(x|z)p(z)}{p(x)},
 \end{equation}
 
 where $p(x)$ is the probability of the data $x$, $p(x|z)$ is the probability of the data given a hypothesis $z$, and $p(z)$ is the probability of that hypothesis $z$. While Bayes theorem by itself can appear non-intuitive or at least difficult to intuit, the key to understanding it is to derive it. It arises directly out of the conditional probability axiom, which itself arises out of the definition of the joint probability. The probability of an event $X$ and an event Y occurring jointly is,

  \begin{equation}
  p(X\cap Y)=p(X|Y)p(Y)
 \end{equation}
 
And since the `AND' is commutative, we have,
 
 \begin{equation}
  p(X\cap Y) = p(Y \cap X) = p(Y|X)p(X)
 \end{equation}
 
 \begin{equation}
  p(X|Y)p(Y)=  p(Y|X)p(X)
  \label{eqn:cond}
 \end{equation}
 
Dividing both sides of Equation~(\ref{eqn:cond}) by $p(Y)$ yields Bayes theorem,
 \begin{equation}
  p(X|Y)=\frac{p(Y|X)p(X)}{p(Y)}
 \end{equation}
 
\begin{table}
  \caption{\textbf{Bayesian Statistics Glossary}}
  \label{tab:bayes_gloss}
  \centering
  \begin{tabular}{l|l}
    Symbol & Name\\\hline
    \hline
$z$ & Latent variable\\\hline
$x$ & Evidence or Data\\\hline
$p(x)$ & Evidence probability\\\hline
$p(z)$ & Prior probability\\\hline
$p(z|x)$ & Posterior probability\\\hline
$p(x|z)$ & Likelihood probability
  \end{tabular}
\end{table}

\section*{Kullback-Leibler Divergence}

When comparing two distributions as we often do in density estimation, the central task of generative models, we need a measure of similarity between both distributions. The Kullback-Leibler divergence is a commonly used similarity measure for this purpose. It is the expectation of the information difference between both distributions. But first, what is information?

To understand what information is and to see its definition, consider the following: The higher the probability of an event, the lower its information content. This makes intuitive sense in that if someone tells us something `obvious' i.e. highly probable i.e. something we and almost everyone else already knew, then that informant has not increased the amount of information we have. Hence the information content of highly probably event is low. Another way to say this is that the information is inversely related to the probability of an event. And since $\log(p(x)$ is directly related to $p(x)$, it follows that $-\log(p(x))$ is inversely related to $p(x)$, and is how we model information:

\begin{equation}
 \mbox{Information content of event x wrt p} = I_p(x)=-\log p(x)
\end{equation}

\begin{equation}
 \mbox{Information content of event x wrt q} =I_q(x)= -\log q(x)
\end{equation}

The difference of information between $q(x)$ and $p(x)$ is therefore:

\begin{equation}
 \Delta I = I_p - I_q = -\log p(x) + \log q(x) = \log\left(\frac{q(x)}{p(x)}\right)
\end{equation}

And the Kullback-Leibler is the expectation of the above difference, and is given by,

\begin{equation}
 D_{KL}(q(x)||p(x)):=E_{\sim q}[ \Delta I] = \int (\Delta I)q(x)  dx = \int q(x) \log\left(\frac{q(x)}{p(x)}\right)dx
 \label{eqn:wrtq}
\end{equation}

Similarly

\begin{equation}
 D_{KL}(p(x)||q(x)):=E_{\sim p}[ \Delta I] = \int (\Delta I)p(x)  dx = \int p(x) \log\left(\frac{p(x)}{q(x)}\right)dx
 \label{eqn:wrtp}
\end{equation}

Note that the Kullback-Leibler (KL) is not symmetric, i.e,

\begin{equation}
 D_{KL}(q(x)||p(x))\neq D_{KL}(p(x)||q(x))
\end{equation}

In $D_{KL}(q(x)||p(x))$, we are taking the expectation of the information difference with respect to $q(x)$ distribution, while in $D_{KL}(p(x)||q(x))$, we are taking the expectation with respect to the $p(x)$ distribution.

Hence the Kullback-Leibler is called a `divergence' and not a `metric' as metrics must be symmetric. There recently have been a number of symmetrization devices proposed for KL which have been shown to improve its generative fidelity~[\cite{puwa2017}][~\cite{chda2017}]~[\cite{arbo2017}].

Note the KL divergence is always non-negative, i.e.,

\begin{equation}
 D_{KL}(q(x)||p(x))=-\int q(x) \log\left(\frac{p(x)}{q(x)}\right)dx \geq 0
\end{equation}

\begin{figure}[h]
\begin{center}
\scalebox{.5}
{\includegraphics{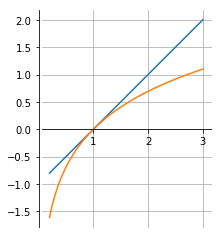}}
\end{center}
\caption[mid]{$\log(t)\leq t-1$}
\label{fig:logx}
\end{figure}

To see this, note that as depicted in Figure~(\ref{fig:logx}), 

\begin{equation}
\log t \leq t - 1
\end{equation}

Therefore

\begin{eqnarray}
  -D_{KL}(q(x)||p(x)) = \int q(x) \log\left(\frac{p(x)}{q(x)}\right)dx \leq\\\nonumber
  \int q(x)\left(\frac{p(x)}{q(x)} - 1\right)dx = \\\nonumber
  \int q(x)\frac{p(x)}{q(x)}dx - \int q(x)dx =\\\nonumber
  \int p(x) dx - \int q(x) dx=\\\nonumber
  1-1=0
\end{eqnarray}

We have just shown,

\begin{equation}
  -D_{KL}(q(x)||p(x))\leq 0
\end{equation}

which implies,

\begin{equation}
 D_{KL}(q(x)||p(x))\geq 0
\end{equation}

\section*{VAE Objective}

\begin{figure}[h]
\begin{center}
\scalebox{.5}
{\includegraphics{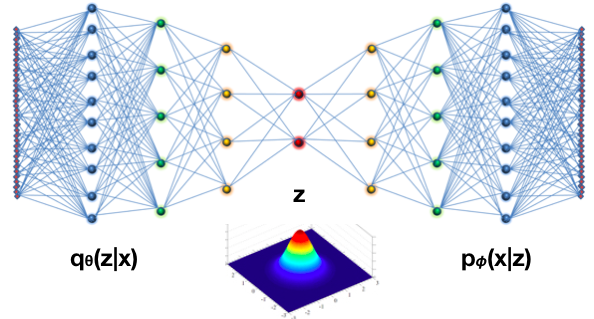}}
\end{center}
\caption[mid]{VAE}
\label{fig:vaeFig}
\end{figure}

Consider variational autoencoders~[\cite{kiwe2013}]. They have many applications including for finer characterization of disease~[\cite{od2019}]. The encoder portion of a VAE yields an approximate posterior distribution $q(z|x)$, and is parametrized on a neural network by weights collectively denoted $\theta$. Hence we more properly write the encoder as $q_{\theta}(z|x)$. Similarly, the decoder portion of the VAE yields a likelihood distribution $p(x|z)$, and is parametrized on a neural network by weights collectively denoted $\phi$. Hence we more properly denote the decoder portion of the VAE as $p_{\phi}(x|z)$. The output of the encoder are parameters of the latent distribution, which is sampled to yield the input into the decoder. A VAE schematic is shown in Figure~(\ref{fig:vaeFig}).

The KL divergence between the approximate and the real posterior distributions is given by,

\begin{equation}
D_{KL}\left(q_{\theta}(z|x_i)||p(z|x_i)\right)=-\int q_{\theta}(z|x_i) \log\left(\frac{p(z|x_i)}{q_{\theta}(z|x_i)}\right)dz  \geq 0
 \end{equation}
 
 Applying Bayes' theorem to the above equation yields,
 
 \begin{equation}
 D_{KL}\left(q_{\theta}(z|x_i)||p(z|x_i)\right)=-\int q_{\theta}(z|x_i) \log\left(\frac{p_{\phi}(x_i|z)p(z)}{q_{\theta}(z|x_i)p(x_i)}\right)dz \geq 0
 \end{equation}
 
 This can be broken down using laws of logarithms, yielding,

  \begin{equation}
 D_{KL}\left(q_{\theta}(z|x_i)||p(z|x_i)\right)=-\int q_{\theta}(z|x_i) \left[\log\left(\frac{p_{\phi}(x_i|z)p(z)}{q_{\theta}(z|x_i)}\right)-\log p(x_i)\right]dz \geq 0
 \end{equation}
 
 Distributing the integrand then yields,
 
   \begin{equation}
-\int q_{\theta}(z|x_i) \log\left(\frac{p_{\phi}(x_i|z)p(z)}{q_{\theta}(z|x_i)}\right)dz + \int q_{\theta}(z|x_i)\log p(x_i)dz\geq 0
 \end{equation}
 
 In the above, we note that $\log(p(x_i))$ is a constant and can therefore be pulled out of the second integral above, yielding,
 
     \begin{equation}
-\int q_{\theta}(z|x_i) \log\left(\frac{p_{\phi}(x_i|z)p(z)}{q_{\theta}(z|x_i)}\right)dz + \log p(x_i)\int q_{\theta}(z|x_i)dz\geq 0
 \end{equation}
 
 And since $q_{\theta}(z|x_i)$ is a probability distribution it integrates to 1 in the above equation, yielding,
 
    \begin{equation}
-\int q_{\theta}(z|x_i) \log\left(\frac{p_{\phi}(x_i|z)p(z)}{q_{\theta}(z|x_i)}\right)dz + \log p(x_i)\geq 0.
 \end{equation}
 
 Then carrying the integral over to the other side of the inequality, we get,
     \begin{equation}
\log p(x_i)\geq \int q_{\theta}(z|x_i) \log\left(\frac{p_{\phi}(x_i|z)p(z)}{q_{\theta}(z|x_i)}\right)dz.
\label{e22}
 \end{equation}
 
 Applying rules of logarithms, we get,

      \begin{equation}
\log p(x_i)\geq \int q_{\theta}(z|x_i) \Big[ \log p_{\phi}(x_i|z) + \log p(z) - \log q_{\theta}(z|x_i)\Big]dz.
 \end{equation}
 
Recognizing the right hand side of the above inequality as Expectation, we write, 
      \begin{equation}
\log p(x_i)\geq E_{\sim q_{\theta}(z|x_i)}\Big[ \log p_{\phi}(x_i|z) + \log p(z) - \log q_{\theta}(z|x_i)\Big]
 \end{equation}

\begin{equation}
\log p(x_i)\geq E_{\sim q_{\theta}(z|x_i)}\Big[ \log p(x_i, z)  - \log q_{\theta}(z|x_i)\Big]
 \end{equation}

 From Equation~(\ref{e22}) it also follows that:
 
      \begin{equation}
\log p(x_i)\geq \int q_{\theta}(z|x_i) \log\left(\frac{p(z)}{q_{\theta}(z|x_i)}\right)dz + \int q_{\theta}(z|x_i) \log p_{\phi}(x_i|z)dz
 \end{equation}
 
       \begin{equation}
\log p(x_i)\geq -D_{KL}(q_{\theta}(z|x_i)||p(z)) + E_{\sim q_{\theta}(z|x_i)}[ \log p_{\phi}(x_i|z)]
\label{elbo}
 \end{equation}
 
The right hand side of the above equation is the Evidence Lower Bound (ELBO) also known as the variational lower bound. It is so termed because it bounds the likelihood of the data which is the term we seek to maximize. Therefore maximizing the ELBO maximizes the log probability of our data by proxy. This is the core idea of variational inference, since maximization of the log probability directly is typically computationally intractable. The Kullback-Leibler term in the ELBO is a regularizer because it is a constraint on the form of the approximate posterior. The second term is called a reconstruction term because it is a measure of the likelihood of the reconstructed data output at the decoder.

Notably, we have some liberty to choose some structure for our latent variables. We can obtain a closed form for the loss function if we choose a gaussian representation for the latent prior $p(z)$ and the approximate posterior, $q_{\theta}(z|x_i)$. In addition to yielding a closed form loss function, the gaussian model enforces a form of regularization in which the approximate posterior have variation or spread (like a gaussian).

\section*{Closed form VAE Loss: Gaussian Latents}
 Say we choose:
 
  \begin{equation}
  p(z)\rightarrow \frac{1}{\sqrt{2\pi\sigma_p^2}}\exp\left(-\frac{(x-\mu_p)^2}{2\sigma_p^2}\right)
 \end{equation}
 
 and
 
   \begin{equation}
  q_{\theta}(z|x_i)\rightarrow \frac{1}{\sqrt{2\pi\sigma_q^2}}\exp\left(-\frac{(x-\mu_q)^2}{2\sigma_q^2}\right)
 \end{equation},
 
 then the KL or regularization term in the ELBO becomes:

 \begin{equation}
  -D_{KL}(q_{\theta}(z|x_i)||p(z))=\nonumber
 \end{equation}

\begin{equation}
\bigintsss \frac{1}{\sqrt{2\pi\sigma_q^2}}\exp\left(-\frac{(x-\mu_q)^2}{2\sigma_q^2}\right) \log \left(\frac{\frac{1}{\sqrt{2\pi\sigma_p^2}}\exp\left(-\frac{(x-\mu_p)^2}{2\sigma_p^2}\right)}{\frac{1}{\sqrt{2\pi\sigma_q^2}}\exp\left(-\frac{(x-\mu_q)^2}{2\sigma_q^2}\right)}\right)dz
\end{equation}

Evaluating the term in the logarithm simplifies the above into,

\begin{eqnarray}
\bigintsss \frac{1}{\sqrt{2\pi\sigma_q^2}}\exp\left(-\frac{(x-\mu_q)^2}{2\sigma_q^2}\right)\times\\\nonumber \left\{-\frac{1}{2}\log(2\pi)-\log(\sigma_p)- \frac{(x-\mu_p)^2}{2\sigma_p^2} + \frac{1}{2}\log(2\pi)+\log(\sigma_q)+ \frac{(x-\mu_q)^2}{2\sigma_q^2}\right\}dz.\nonumber
\end{eqnarray}

This further simplifies into,

\begin{eqnarray}
\frac{1}{\sqrt{2\pi\sigma_q^2}}\bigintsss \exp\left(-\frac{(x-\mu_q)^2}{2\sigma_q^2}\right)\left\{-\log(\sigma_p)- \frac{(x-\mu_p)^2}{2\sigma_p^2} + \log(\sigma_q)+ \frac{(x-\mu_q)^2}{2\sigma_q^2}\right\}dz,
\end{eqnarray}

which further simplifies into,
\begin{eqnarray}
\frac{1}{\sqrt{2\pi\sigma_q^2}}\bigintsss \exp\left(-\frac{(x-\mu_q)^2}{2\sigma_q^2}\right)\left\{\log\left(\frac{\sigma_q}{\sigma_p}\right)- \frac{(x-\mu_p)^2}{2\sigma_p^2} + \frac{(x-\mu_q)^2}{2\sigma_q^2}\right\}dz.
\end{eqnarray}

Expressing the above as an Expectation we get,

\begin{eqnarray}
 -D_{KL}(q_{\theta}(z|x_i)||p(z))=E_q\left\{\log\left(\frac{\sigma_q}{\sigma_p}\right)- \frac{(x-\mu_p)^2}{2\sigma_p^2} + \frac{(x-\mu_q)^2}{2\sigma_q^2}\right\}\\\nonumber
 =\log\left(\frac{\sigma_q}{\sigma_p}\right)+ E_q\left\{-\frac{(x-\mu_p)^2}{2\sigma_p^2} + \frac{(x-\mu_q)^2}{2\sigma_q^2}\right\}\\\nonumber
 =\log\left(\frac{\sigma_q}{\sigma_p}\right)- \frac{1}{2\sigma_p^2}E_q\left\{(x-\mu_p)^2\right\} + \frac{1}{2\sigma_q^2}E_q\left\{(x-\mu_q)^2\right\}\\\nonumber
\end{eqnarray}

And since the variance $\sigma^2$ is the expectation of the squared distance from the mean, i.e., 

\begin{equation}
 \sigma_q^2=E_q\left\{(x-\mu_q)^2\right\},
\end{equation}

it follows that,

\begin{eqnarray}
 -D_{KL}(q_{\theta}(z|x_i)||p(z))  =\log\left(\frac{\sigma_q}{\sigma_p}\right)- \frac{1}{2\sigma_p^2}E_q\left\{(x-\mu_p)^2\right\} + \frac{\sigma_q^2}{2\sigma_q^2}\\\nonumber
  =\log\left(\frac{\sigma_q}{\sigma_p}\right)- \frac{1}{2\sigma_p^2}E_q\left\{(x-\mu_p)^2\right\} + \frac{1}{2}\\\nonumber
 =\log\left(\frac{\sigma_q}{\sigma_p}\right)- \frac{1}{2\sigma_p^2}E_q\left\{(x-\mu_q+\mu_q-\mu_p)^2\right\} + \frac{1}{2}\\\nonumber
  =\log\left(\frac{\sigma_q}{\sigma_p}\right)- \frac{1}{2\sigma_p^2}E_q\left\{(\underbrace{x-\mu_q}_a+\underbrace{\mu_q-\mu_p}_b)^2\right\} + \frac{1}{2}\\\nonumber
\end{eqnarray}

Recall that,

\begin{equation}
 (a+b)^2 = a^2+2ab+b^2,
\end{equation}

therefore,

\begin{eqnarray}
 -D_{KL}(q_{\theta}(z|x_i)||p(z))=\log\left(\frac{\sigma_q}{\sigma_p}\right)- \frac{1}{2\sigma_p^2}E_q\left\{(\underbrace{x-\mu_q}_a+\underbrace{\mu_q-\mu_p}_b)^2\right\} + \frac{1}{2}\\\nonumber
=\log\left(\frac{\sigma_q}{\sigma_p}\right)- \frac{1}{2\sigma_p^2}E_q\left\{(x-\mu_q)^2+2(x-\mu_q)(\mu_q-\mu_p)+(\mu_q-\mu_p)^2\right\} + \frac{1}{2}\\\nonumber
=\log\left(\frac{\sigma_q}{\sigma_p}\right)- \frac{1}{2\sigma_p^2}E_q\left\{(x-\mu_q)^2+2(x-\mu_q)(\mu_q-\mu_p)+(\mu_q-\mu_p)^2\right\} + \frac{1}{2}\\\nonumber
=\log\left(\frac{\sigma_q}{\sigma_p}\right)- \frac{1}{2\sigma_p^2}\left[E_q\left\{(x-\mu_q)^2\right\}+2E_q\left\{(x-\mu_q)(\mu_q-\mu_p)\right\}+E_q\left\{(\mu_q-\mu_p)^2\right\}\right] + \frac{1}{2}\\\nonumber
=\log\left(\frac{\sigma_q}{\sigma_p}\right)- \frac{1}{2\sigma_p^2}\left[\sigma_q^2+2*0*(\mu_q-\mu_p)+(\mu_q-\mu_p)^2\right] + \frac{1}{2}\\\nonumber
=\log\left(\frac{\sigma_q}{\sigma_p}\right)- \frac{\sigma_q^2+(\mu_q-\mu_p)^2}{2\sigma_p^2} + \frac{1}{2}\\\nonumber
\end{eqnarray}

And when we take $\sigma_p=1$ and $\mu_p=0$, we get,

\begin{eqnarray}
  -D_{KL}(q_{\theta}(z|x_i)||p(z))=\log\left(\sigma_q\right)- \frac{\sigma_q^2+\mu_q^2}{2} + \frac{1}{2}\\\nonumber
=\frac{1}{2}\log\left(\sigma_q^2\right)- \frac{\sigma_q^2+\mu_q^2}{2} + \frac{1}{2}\\\nonumber
=\frac{1}{2}\Bigg[1+ \log\left(\sigma_q^2\right)- \sigma_q^2-\mu_q^2\Bigg]
\end{eqnarray}

Recall the ELBO, Equation~(\ref{elbo}),

\begin{equation}
\log p(x_i)\geq -D_{KL}(q_{\theta}(z|x_i)||p(z)) + E_{\sim q_{\theta}(z|x_i)}\Big[ \log p_{\phi}(x_i|z)\Big]\nonumber
 \end{equation}
 
 From which it follows that the contribution from a given datum $x_i$ and a single stochastic draw towards the objective to be \textit{maximized} is,
 
        \begin{equation}
\frac{1}{2}\Bigg[1+ \log\left(\sigma_j^2\right)- \sigma_j^2-\mu_j^2\Bigg] + E_{\sim q_{\theta}(z|x_i)}\Big[ \log p_{\phi}(x_i|z)\Big]
 \end{equation}

 where $\sigma_j^2$ and $\mu_j$ are parameters into the approximate distribution, $q$, and $j$ is an index into the latent vector $z$. For a batch, the objective function is therefore given by,

\begin{equation}
{\cal{G}}=\sum_{j=1}^J\frac{1}{2}\Bigg[1+ \log\left(\sigma_i^2\right)- \sigma_i^2-\mu_i^2\Bigg] + \frac{1}{L}\sum_lE_{\sim q_{\theta}(z|x_i)}\Big[ \log p(x_i|z^{(i,l)})\Big]
\label{eqn:gain}
 \end{equation}
 
 where $J$ is the dimension of the latent vector $z$, and $L$ is the number of samples stochastically drawn according to re-parametrization trick.
 
 Because the objective function we obtain in Equation~(\ref{eqn:gain}) is to be maximized during training, we can think of it as a `gain' function as opposed to a loss function. To obtain the loss function, we simply take the negative of ${\cal{G}}$:
 
 \begin{equation}
{\cal{L}}=-\sum_{j=1}^J\frac{1}{2}\Bigg[1+ \log\left(\sigma_i^2\right)- \sigma_i^2-\mu_i^2\Bigg] - \frac{1}{L}\sum_lE_{\sim q_{\theta}(z|x_i)}\Big[ \log p(x_i|z^{(i,l)})\Big]
\label{eqn:loss}
 \end{equation}
 
 Therefore to train the VAE is to seek the optimal network parameters $(\theta^*, \phi^*)$ that minimize ${\cal{L}}$:
 
 \begin{equation}
 (\theta^*, \phi^*)=argmin_{(\theta, \phi)} {\cal{L}}(\theta,\phi)
 \end{equation}

\section*{Conclusion}

We have done a step-by-step derivation of the VAE loss function. We illustrated the essence of variational inference along the way, and have derived the closed form loss in the special case of gaussian latent.

\section*{Acknowledgement}

The author thanks Larry Carin for helpful discussion on consequences of Kullback-Leibler divergence asymmetry, and on KL symmetrization approach.

\end{document}